# Development of a classifiers/quantifiers dictionary towards French-Japanese MT

**Mutsuko Tomokiyo**  Mutsuko.Tomokiyo@imag.fr
**Mathieu Mangeot**  Mathieu.Mangeot@imag.fr
**Christian Boitet**  Christian.Boitet@imag.fr
UGA, GETALP-LIG, bâtiment IMAG, 700 avenue Centrale,
Domaine Universitaire de Saint-Martin-d'Hères, CS 40700
38058 Grenoble CEDEX 9, France

**Abstract**

Although classifiers/quantifiers (CQs) expressions appear frequently in everyday communications or written documents, they are described neither in classical bilingual paper dictionaries, nor in machine-readable dictionaries. The paper describes a CQs dictionary, edited from the corpus we have annotated, and its usage in the framework of French-Japanese machine translation (MT).

CQs treatment in MT often causes problems of lexical ambiguity, polylexical phrase recognition difficulties in analysis and doubtful output in transfer-generation, in particular for distant languages pairs like French and Japanese.

Our basic treatment of CQs is to annotate the corpus by UNL-UWs (Universal Networking Language - Universal words)[1], and then to produce a bilingual or multilingual dictionary of CQs, based on synonymy through identity of UWs.

**Keywords**: classifiers, quantifiers, corpus annotation, UNL, UWs dictionary, phraseology study, Tori Bank, French-Japanese MT

## Introduction

We call CQs (classifiers/quantifiers) words or phrases which are used in some languages to indicate the class of a noun or a nominal phrase, depending upon the type of its referent or upon speaker's observation of the referent, when they appear in quantitative expressions. They denote:

(a) CQs expressing quantity of the referent by counting.
Eg. pièce (piece) (in French), 枚(mai, sheet), 点 (ten, piece) (in Japanese), cm, gram

(b) CQs representing quantity concept, based on speaker's observation or general metonymy.
Eg. un brin de (a little), bribes de (scraps of), ひとつまみの (hito-tsumami no, a pinch of), 山盛りの (yama-mori no, a pile of).

There are two cases for a CQ: (1) it can belong to only the (a) type or the (b) type, and (2) it can belong at the same time to both the (a) and (b) types. That is because, on the one hand, there are some CQs that play only the role of classifier or quantifier, and, on the other hand, there are CQs that play both of these roles.

---

[1] The UNL (Universal Networking Language) project was founded at the Institute of Advanced Studies (IAS) of the United Nations University in Tokyo in April 1996 under the aegis of UNU (United Nations University, Tokyo) and with financial support from ASCII corporation (a Japanese publishing company, 1977-2002) and UNL-IAS. http://www.undl.org/unlsys/unl/unl2005/attribute.htm

Eg. un brin de paille (a wisp of straw), un brin de folie (a touch of madness)[2].

When we started to deal with CQs expressions in the framework of French-Japanese MT, we met mainly the following difficulties, which were inherent in QCs:

1. Resolution of lexical ambiguity of polysemic nouns
   Eg. pièce (piece) : (Japanese translation as CQs) 枚(mai, sheet or $\phi$[3]), 点 (ten, $\phi$), 頭(tou, $\phi$), 樽(taru, cask), etc.

2. Producing adequate CQs in Japanese when they are absent in French
   Eg. deux livres (two books) : (Japanese translation) 二冊の本 (ni-satsu no hon)

   ni = two, satsu = $\phi$, no = postposition, hon = book, where 冊 (satsu) is one of the Japanese CQs for books, notebooks, albums, etc.

3. Normalization for floating quantifier phenomenon in Japanese

4. Recognition of QC polylexical expressions over the course of corpus development
   Eg. une pincée de sel (a pinch of salt): (Japanese translation) ひとつまみの塩 (hito-tsumami no shio)

   hito = 1, tsumami = pinch, no = of , shio = salt

To handle these linguistic behaviours of CQs in a comprehensive manner, we have adopted the UNL-UWs format for our corpus annotations and dictionary descriptions. Another motivation is the desire to be able to extend this work to many other languages, in the framework of MT based on the passage through the UNL semantic pivot.

In this paper, we first examine the behaviour of CQs and the related problematic issues more concretely, from the point of view of French↔Japanese MT, and then propose a resolution of the above-mentioned problems by extending the UNL-UWs dictionary.

## 1 Lexical ambiguity for classifiers/quantifiers

According to our studies on ambiguities for MT, 14% of analysis errors are due to polysemous words[4] [Boitet and Tomokiyo (1995), Boitet and Tomokiyo (1996), Tomokiyo and Axtmeyer (1996)]. Also, Wisniewski et al. (2013) say the most frequent necessary post-editing operation in their French corpus translation into English is to correct articles like "les", "le", "du", etc., and the next one concerns lexical transfer errors of polysemous words.

We have also confirmed that, when polysemous words are used in their abstract or figurative meaning in CQs expressions, translation results produced by current MT systems are not at all good, because words contained in CQ phrases are often at the same time polysemous and are used in their figurative meaning.

The following example shows « pincée (pinch, つまみ, tsumami) » appearing in a quantifier phrase « une pincée de », and used in its figurative meaning. When one looks at the translation outputs produced by free as well as commercial MT systems, it appears that there is a lack of phraseology studies and polysemy disambiguation method for the word « pincée »[5].

---

[2]"brin" means (1) a small stalk, and (2) "a bit, a little" in "un brin de"

[3]The symbol $\phi$ means the absence of corresponding translation in French.

[4]We have carried on a research on ambiguity analysis from the lexical, semantic and contextual points of view since 1996. Ambiguities have been defined, categorized, and formalized as objects in an ambiguity database, and we have used this theoretical background to label ambiguities in Japanese-English interpreted dialogues, collected for the development of a speech translation system at ATR in Japan (1996 ).

[5]The word "pincée" is used as CQs in form of "une pincée de"+noun without particle" for pulverized substances.

Table 1: problem of CQ words ambiguity in French-Japanese MT

| French word | Examples[7] | English translation | Japanese translation |
|---|---|---|---|
| pièce | une pièce de toile | a piece of cloth | 一枚 (ichi-mai) の布 |
| | une pièce de mobilier | a piece of furniture | 一点 (it-ten) の家具 |
| | dix pièces de bétail[8] | ten pieces of cattle | １０種 (jyut-tou) の家畜 |
| | plusieurs pièces de bois | several pieces of wood | 数枚 (suu-mai) の板 |
| | Une pièce de vin est un tonneau de vin contenant environ 220 litres. | A cask of wine is a barrel of wine containing about 220 liters. | 一樽 (hito-taru) のワインとは約２２０リットルを含むワイン樽である。 |
| | J'ai reçu une demi-pièce de ce vin. | I received half a cask of this wine. | わたしは半樽(han-taru) のワインを受け取った。 |
| | Dans une pièce de théâtre, il n'y a pas de narrateur pour raconter les faits. | In a play, there is no narrator to tell the facts. | ある作品 (aru-sakuhin) では事実を語るナレータが いない。 |
| | une pièce de viande | a piece of meat | 一切れの肉 (hito-kire) |
| | une pièce de blé | a wheat field | 一枚 (ichi-mai) の麦畑 (no mugi-batake) |

Eg. Ajoutez une pincée de sel. （ひとつまみの塩を加えなさい (hitotsumami-no shio-wo kuwaenasai), Add( 加えなさい) a pinch of (ひとつまみの) salt (塩).) → (translation outputs) 塩のつねり (tsuneri) を加えなさい / 塩のピンチ (pinchi) を加えなさい / 塩のピンチ (pinchi) を追加します (shio no tsuneri wo kuwaenasai / shio no pinchi wo kuwaenasai / shio no pinchi wo tsuikashimasu)[6].

Even measure words like cm, km, kg, etc. have acronym ambiguity [Mari (2011)].
Eg. cm ← centimètre, congrégation de la mission, coût marginal, etc.

To disambiguate a polysemic CQ, we describe each of its meanings, with the associated conditions of occurrence, as a UW (contained in our Universal Words dictionary).

In our fr-UW dictionary, the description for the ambiguous word "pièce" is as follows:
pièce → cask(icl>wine)
pièce → piece(icl>cloth)
pièce → piece(icl>furniture)
pièce → piece(icl>meat)
pièce → room(icl>place)

---

[6]The translations on following MT systems don't make sense.
http://www.reverso.net/translationresults.aspx?lanḡFR&direction̄francais-japonais.
http://www.worldlingo.com/fr/products_services/worldlingo_translator.html. https://translate.google.com/#fr/en/a

[7]The sources of the examples are the French-Japanese dictionary "Royal", the information on "pièce" in the Wiktionary "Vinothèque" article, see https://fr.wiktionary.org/wiki/pièce_de_vin, and http://www.etudes-litteraires.com/etudier-piece-de-theatre.php

[8]Each animal, like ox, cow, etc., that belongs to cattle. One says rather "head of cattle" today.

[9]The actant means here an expression that helps complete the meaning of a predicate.

[10]The semantic relation labels are created from UNL ontology, which store all relational information in a lattice structure, where UWs are interconnected through relations including hierarchical relations (10 levels) such as "icl" (a-kind-of) and "iof" (an-instance-of), and mean headword's sub-meaning and equivalent quantity, respectively. http://www.undl.org/unlexp/

Table 2: UWs and UWs dictionary

---

A UW is a character string of the form "headword(constraint_list)" which represents a concept associated to the headword. For example, "look(agt>thing, equ>search, icl>examine(icl>do, obj>thing))" is a possible UW for the meaning of the verb "look" corresponding to "examine". Other UWs will be used for various meanings of "look" as a noun: appearance (Paul's look(s)), or action (after a quick look,...).

The semantic representation of an utterance in UNL is a hypergraph, where each node bears a UW, possibly augmented by semantic attributes, and arcs bear semantic relations from a small list of about 40, like "agt", "obj", "aoj", "ben".

In fact, there are three types of UW: *r*estricted UWs, which are formed as said above (headword plus constraint list), *e*xtra UWs, which are a special type of restricted UWs, and *b*asic UWs, which are bare headwords, with no constraint list.

The syntax for dictionary description is:

```
<UW> ::= <Headword>['('<Constraint_List>')']
```

The constraint list restricts the interpretation of a UW to a specific concept included within those covered by the Basic UW [Uchida et al. (2006)], or to a subset of them. Eg.
    look(agt>thing, equ>search, icl>examine(icl>do, obj>thing))
    relever (to season): season(agt>person, obj>dish, icl>action)
    樽 (taru, to cask): cask(icl>wine, equ>220 litres)

The semantic relation "agt" denotes that the first actant[9] of "look" is a "thing", "look" belongs to equivalent semantic level in UNL ontology map[10] with "search", and includes the meaning of "examine", "examine" is an action verb and its grammatical object is a noun meaning things.

The UNL-lang dictionaries contained, at the moment of writing, 1269421 headwords for Japanese, 520305 headwords for French and 1458686 headwords for English. The semantic attributes consist of 58 labels and semantic relation labels [Uchida et al. (2006)].

For French-Japanese translation, French words are converted into UWs by using a UNL-French dictionary, and a UNL-Japanese dictionary is used for generating Japanese translations.

## 2 Handling dummy classifiers

A frequent but difficult case appears when a CQ does not appear explicitly in one language of a source-target language pair[11], nevertheless they are mandatory in type (a) CQ usage, like 冊 (satsu) for counting books, notebooks, albums, etc., 匹 (hiki) for counting small animals, 台 (dai) for counting cars, bicycles, pianos, computers, etc. Eg.

2 livres (two books) → 二冊の本 (ni-satsu no hon)
    ni = 2, satsu = $\phi$, no = $\phi$, hon = books
un chat (a cat) → 一匹の猫 (i-ppiki no neko)
    i = 1, ppiki = $\phi$, no = $\phi$, neko = cat

There is no lexeme in French corresponding to 冊 (satsu), but if 冊 (satsu) is omitted in the translation into Japanese, the sentence doesn't make sense. In order to represent such Japanese sentences in UNL, which is based on English, when these CQs don't exist in English, we create new UWs beginning by "CQ-<romanized Japanese CQ>", followed by a list of some English referent nouns. For example: CQ-satsu-books-notebooks-albums, "CQ-dai-cars-bicycles-computers-pianos"[12].

Absent CQs in French are marked by the attribute "@eld" (elided), which we have added to the original attribute list.

Eg. Description for 冊 (satsu) in Japanese-UW dictionary:
冊 (satsu) (icl>CQ-books, notebooks, albums)
Accordingly, the graphs for 二冊の本 (two books) is as follows:
qua(book(icl>thing).@pl, :01)
mod:01(CQ-satsu-books-notebooks-albums(icl>CQ).@entry.@eld, 2)

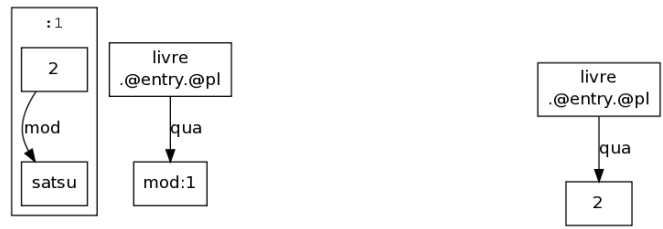

(a) Tentative japanized UNL-graph for "二冊の本 (two books)"

(b) Tentative frenchized UNL-graph for "deux livres (two books)"

---

[11]This happens not only between Japanese and western languages, but also between French and English: eg. une pièce de blé → a wheat field, une pièce de théâtre → a play

[12]At present, new CQs are made by indicating only some modifiable nouns, but this should be completed by labels coming from Mel'chuk's labels in the "Dictionnaire explicatif et combinatoire du français contemporain (DEC)" (1999, Montréal, UdM Press). In the DEC, a word is analyzed from 5 points of view: general morphosyntax, semantics, syntactic combinatorics, lexical co-occurence, phraseology. The analysis of the lexical co-occurences is made by using 60 labels corresponding to as many lexico-semantic functions (FLs) such as Magn, Anti-Magn, Mult, Sing, etc. Magn(X) is "very X", Mult(X) is "a regular quantity of X" and Sing(X) is "a regular quantum of X".

Values of FLs are subsets of lexemes, ordered by degree of intensity of the relation. For example, Magn(fever) = {high, strong, horse}, Mult(fish) = {shoal, school}, and Sing(wine) = {glass, bottle, cask, liter...}.

When possible, we will use these labels instead of the above labels such as "CQ-concrete nouns". Note that it is not possible in cases where two or more Japanese counters corresponding to different measures can apply to the same nominal concept, but don't exist in English: to use only the FL label would lead to a loss of information and to the impossibility of exact translation. Examples:
CQ-tou = [qua(mod(icl>animal, Magn), number]
CQ-piki = [qua(mod(icl>animal, Anti-Magn), number]

Table 3: Positions of numerical phrases in Japanese

| Morphology | Japanese sentence and English translation | words order and word-to-word correspondance to English translation |
|---|---|---|
| Numerical word and CQs | 本を二冊買いました (I bought two books.) | hon = book, wo = postposition($\phi$), ni = 2, satsu= $\phi$, kaimashita = bought |
| Numerical word +CQs+の(no, of)+Noun | 二冊の本を買いました(I bought two books.) | ni=2, satsu=$\phi$, no = postposition($\phi$), hon = books, wo = postposition($\phi$) kaimashita = bought |
| Noun+Numerical word+CQs | 本二冊買いました(I bought two books.) | hon = books, ni = 2, satsu = $\phi$, kaimashita = bought |
| Numerical word+CQs | 本を買いました，二冊 (I bought books, two.) | hon = books, wo = postposition($\phi$), kaimashita = bought, ,=comma, ni= 2, satsu = $\phi$ |

## 3  Association of numerical phrase with its host phrase

There are two different aspects concerning the floating quantifier behaviour in Japanese [Miyagawa (1989)].

Firstly, the problem we have encountered in the process of Japanese-French MT, lies in the fact that the Japanese quantifiers can be freely positioned among phrase units in a sentence.

The "Numerical word + CQ + の (no, of) + Noun" type can be split into the CQ phrase and the «Noun» part, in which case a CQ phrase behaves like an adverb before the predicative verb in a sentence. Hence, three types of expressions are possible for the same meaning [Miyagawa (1989)].

Standardization of a floating CQ position consists in determining the CQ phrase and its host phrase, when they are separated in a sentence. In fact, the floating quantifier phenomenon exists also in French, although its linguistic behaviour is different[13] from the Japanese case. Hence, we need modifiable nouns information for each quantifier in order to find out their host noun phrase.

Secondly, there is a risk of generating meaningless expressions as a Japanese translation outputs in some cases, when the association condition between a floating CQ and its host phrase is not given. For instance, "3kgの子豚がいました" (3kg-no kobuta-ga imashita) (There was a 3kg piglet.) is acceptable as a Japanese sentence, but "子豚が3kgいました" (kobuta-ga 3kg imashita)*[14] doesn't make sense, because «子豚 (kobuta, piglet)» means only an alive piglet and co-occurs with "いました" (there was), but "3kg" cannot[15]. Hence, to avoid a machine translation output such as "子豚が3kgいました" (observed), supplementary information on "子豚" on the verb "いる" (iru, there is, or exists) and on how to use that information is necessary. For that reason, we also use a UNL-jp dictionary, which enables us to describe semantic cooccurence information between words (here, japanese lemmas).

In order to find the host phrase of a floating CQ, that is, to get the same translation results for the sentences which are morphologically different but have the same meaning,

---

[13]Floating CQs in French are "tous", "toutes", etc., number and gender agreement is obligatory between two phrases [Miyagawa (1989), Bobaljik (2001)], whereas there are neither number nor gender for common nouns in Japanese.

[14]子豚が3kgいました*, For the piglet, there were 3 kg*.

[15]There are two verbs expressing "existence" or "presence" in Japanese: "いる (iru)" for human being and animals and "ある" (aru) for things

we add some information to "aoj", mentioned above in the square.
Descriptions for いる (iru) and ある (aru) are as follows.
いる (iru) : there-be(obj>animal)
いる (iru) : there-be(obj>person)
ある (aru) : there-be(obj>thing)

## 4 Recognition of quantifiers/classifiers and phraseology

The Type (a) CQs above-mentioned come from Phrase Book II, Tori Bank[16] (see Annex 1), while referring to existing weights and measures dictionaries[17]. Phrases book II includes basic CQs which were manually or semi-automatically collected from journals, novels, numerous articles on the Web, etc. in French and Japanese[18]. To extend this, we are using the "Cesselin" Japanese-French dictionary[19] and the "Tangorin" Japanese-English dictionary[20], in which we have annotated some headwords as potential CQs, according to originally given indications[21]. For the Type (b) one, it's laborious to pin down phrasemes[22] in row data. Eg.
une poignée de sable (a handful of sand), une pointe d'ironie (a touch of irony), un pouce de terre (a handful of soil).

French and English phrasemes are, however in many cases, composed of "Number + Noun + preposition (de, of) + Noun without article".

The Type (b) CQs in the Phrase Book II have been collected from a parallel corpus according to the frequency of polylexical expressions, by using a software that can produce a list of keywords in context [23]. We have filtered the collected data as CQs by checking them with the UWs in the dictionary.

## 5 Specification of classifiers/quantifiers dictionary

We anticipate that our CQs dictionary will include about 8000 entries for each language according to manual count by 1% (8269 entries) random sampling from the Cesselin dictionary (its total number of entries is 826970).

At present, our CQs dictionary contains 3000 entries. The specification (microstructure) of its entries is as follows:

---

[16]Tori Bank is a sentence corpus which has developed at Tottori Unversity in Japan in 2007. http://unicorn.ike.tottori-u.ac.jp/toribank/about_toribank.html

[17]Cassell's French-English, English-French dictionary: with appendices of proper names, French coins, weights, and measures with conversion tables.

[18]At present, the total number of registered entries is about 2000 for the Type (a) CQs and 1000 for Type (b) CQs, and it is becoming larger day by day.)

[19]The Cesselin is a printed dictionary published in 1939 and 1957 in Japan. It has been reprocessed into a numeric version equipped with a search engine by Mathieu Mangeot-Nagata in 2015 [Mangeot-Nagata (2016)]: https://jibiki.fr).

[20]http://tangorin.com/

[21]Eg. ken (軒) in the Cesselin (English translations have been added by us.)
ken (軒) n.m. Avant-toit, f. Maison. spé: s'emploie pour compter les maisons (special: used to count houses).
十二軒 (Jyû ni ken, 12 houses) douze maisons, 二 軒目です (C'est la deuxième maison, It's the second house)
けん ken 軒 in the Tangorin dictionary:
suffix / counter:
1. counter for buildings (esp. houses)
彼女は鳥かごを軒からつるした。She hung the cage from the eaves.
彼の叔父は家を十軒も持っている。His uncle owns no fewer than ten houses.

[22]By "phraseme" we mean a set phrase, an idiomatic phrase, a polylexical expression, etc.

[23]http://en.wikipedia.org/wiki/Sketch_Engine

Table 4: Type (b) CQs in Phrase Book II : "pointe"

| French word | examples | Source | Japanese translation | English translation |
|---|---|---|---|---|
| Pointe | une pointe d'ironie mal placée | J.L.Carré | 場違いの皮肉をちくりと | the tip of , a hint of, a note of, a trace of |
|  | relever la sauce avec une pointe d'ail | Livre de cuisine | ソースにニンニクをちょっときかせる | pick up the sauce with a hint of garlic |
|  | avec une pointe d'agacement dans la voix | T.Jonquet | 声にすこし苦しみをにじませて | with a hint of irritation in the voice |

Table 5: KWIC of "pointe" from Sketch Engine

| doc#357 | qui marque le déclin définitif de cette | pointe | de poussée et de sécrétions des hormones |
|---|---|---|---|
| doc#397 | la sierra Pacaraima, qui constituent une | pointe | avancée du Sertao brésilien. </p><p> En janvier |
| doc#457 | de nouveauté, un soupçon de douceur, une | pointe | d'exotisme : commence par te mettre dans |
| doc#517 | Tafer ne sont capables d'évoluer seuls en | pointe | . </p><p> Arles - Marseille En concédant une |

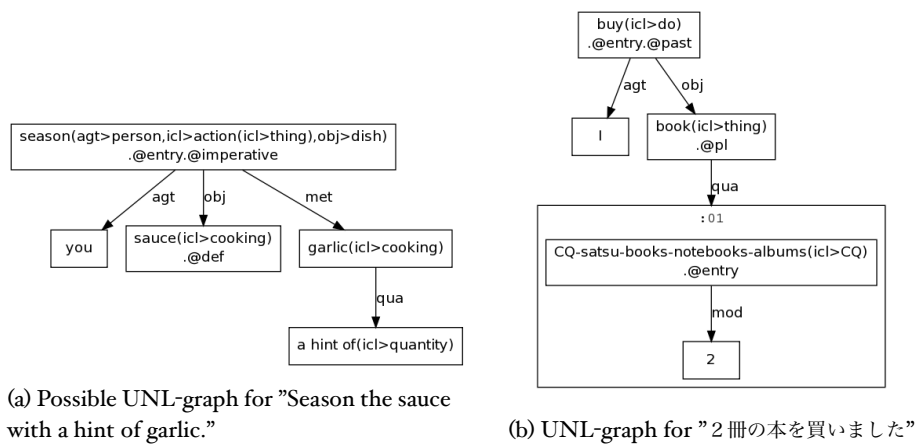

(a) Possible UNL-graph for "Season the sauce with a hint of garlic."

(b) UNL-graph for "2冊の本を買いました"

Figure 2: Two UNL-graphs representing sentences containing CQs

Table 6: Description of "pointe"

| items | description for "pointe" |
|---|---|
| 1.Identification number | XX |
| 2. Keywords and class | pointe (n.) |
| 3. English sentence | Season the sauce with a hint of garlic |
| 4. French sentence | relever la sauce avec une pointe d'ail |
| 5. Japanese sentence | ソースにニンニクをちょっときかせる |
| 6. Source | Royal |
| 7. UNL annotation | agt(season(agt>person, obj>dish, icl>action>thing).@entry.@imperative, you)<br>obj(season(agt>person, obj>dish, icl>action>thing).@entry.@imperative, sauce(icl>cooking).@def)<br>met(season(agt>person, obj>dish, icl>action>thing).@entry.@imperative, garlic(icl>cooking))<br>qua(garlic(icl>cooking), a hint of(icl>quantity)) |

**Perspectives and Conclusion**

We have studied the methodology for phraseology treatment on MT systems, while developing a French-Japanese-English parallel corpus and have known deeper linguistic analysis [Petit (2004), Gouverneur (2005)] is necessary for CQs dictionary description.

The corpus will be made freely accessible, so that software developers can use it. It should also be helpful for learners of languages, because it covers lexico-semantic information which cannot yet be found in any bilingual dictionary. We intend to produce a tool bilingual sentence-aligned corpus processing tool that will show corresponding (chunks of) words between 2 languages are shown on demand by character blinking or where the meaning of nouns or verbs in a sentence is shown without any ambiguity by interpreting UNL annotations. A prototype has been already presented by a Ph.D student in his thesis [Chenon (2005)].

Table 7: Description of 冊

| items | description for "冊" |
|---|---|
| 1. Identification number | XX |
| 2. Keywords and class | satsu(CQ-books, notebooks, albums) |
| 3. English sentence | I bought 2 books. |
| 4. French sentence | J'ai acheté 2 livres. |
| 5. Japanese sentence | "2冊の本を買いました。" |
| 6. Source | Royal |
| 7. UNL annotation | agt(buy(icl>do).@entry.@past, I)<br>obj(buy(icl>do).@entry.@past, book(icl>thing).@pl)<br>qua(book(icl>thing).@pl, :01)<br>mod:01(CQ-satsu-books-notebooks-albums(icl>CQ).@entry.@eld, 2) |

**Annex** 「鳥バンク」

(examples from the Tori-Bank)
     Eg. 「塁 (rui, base)」, 「寸 (sun, approx. 3.03 cm)」
AC00046100 P11:二塁走者の生還を許し:VP@28:allowing the runner to score from second:VP
AC00046100 P4:一塁へ悪投し、:VP@7:threw wild to first:VP
AC01599600 C6:一寸先も見え:CL@27:we could not see an inch ahea:CL
AC01599600 P6:一寸先も見え:VP@40:see an inch ahead:VP